\documentclass[a4paper,twoside]{article}

\usepackage{epsfig}
\usepackage{subcaption}
\usepackage{calc}
\usepackage{amssymb}
\usepackage{amstext}
\usepackage{amsmath}
\usepackage{amsthm}
\usepackage{multicol}
\usepackage{multirow}
\usepackage{pslatex}
\usepackage{gensymb}
\usepackage{apalike}
\usepackage{algorithm2e}
\usepackage[bottom]{footmisc}
\usepackage{xcolor}
\usepackage{graphicx}
\usepackage{SCITEPRESS}  
\usepackage{url}

\begin{document}

\title{Detecting 3D Line Segments for 6DoF Pose Estimation with Limited Data}

\author{
\authorname{Matej Mok\sup{1}\orcidAuthor{0009-0006-4737-0799}, Lukáš Gajdošech\sup{1}\sup{2}\orcidAuthor{0000-0002-8646-2147}, Michal Mesároš\sup{2}\orcidAuthor{0009-0005-7048-8341}, Martin Madaras\sup{1}\sup{2}\orcidAuthor{0000-0003-3917-4510}, Viktor Kocur \sup{1}\orcidAuthor{0000-0001-8752-2685
}}
\affiliation{\sup{1}Faculty of Mathematics, Physics and Informatics, Comenius University Bratislava, Slovakia}
\affiliation{\sup{2}Skeletex Research, Bratislava, Slovakia}
\email{\{matej.mok, lukas.gajdosech, viktor.kocur\}@fmph.uniba.sk, \{mesaros, madaras\}@skeletex.xyz}
}


\abstract{
The task of 6DoF object pose estimation is one of the fundamental problems of 3D vision with many practical applications such as industrial automation. Traditional deep learning approaches for this task often require extensive training data or CAD models, limiting their application in real-world industrial settings where data is scarce and object instances vary. We propose a novel method for 6DoF pose estimation focused specifically on bins used in industrial settings.
We exploit the cuboid geometry of bins by first detecting intermediate 3D line segments corresponding to their top edges. Our approach extends the 2D line segment detection network LeTR to operate on structured point cloud data. The detected 3D line segments are then processed using a simple geometric procedure to robustly determine the bin's 6DoF pose. To evaluate our method, we extend an existing dataset with a newly collected and annotated dataset, which we make publicly available. We show that incorporating synthetic training data significantly improves pose estimation accuracy on real scans. Moreover, we show that our method significantly outperforms current state-of-the-art 6DoF pose estimation methods in terms of the pose accuracy (3 cm translation error, 8.2$^\circ$ rotation error) while not requiring instance-specific CAD models during inference.}

\onecolumn 
\maketitle \normalsize 
\setcounter{footnote}{0} \vfill

\begin{figure*}[h]
\centering
\includegraphics[width=\textwidth]{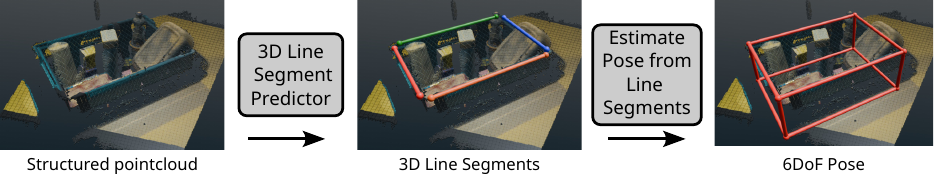}
\caption{Our 6DoF bin pose estimation pipeline from structured point cloud. To estimate the 6DoF pose we first detect 4 3D line segments. A robust geometric regression scheme is then used to obtain the final pose.}
\label{pipeline} 
\end{figure*}

\section{\uppercase{Introduction}}
\label{sec:introduction}

Automating industrial processes can significantly boost efficiency in various areas of industry. Computer vision is often an important aspect of such automation efforts as it enables the processing of visual data from cameras or other sensors. One such application of computer vision is the localization of objects in 3D space using 3D cameras. In this paper, we deal specifically with the problem of estimating the poses of bins captured by 3D cameras. When the pose is accurately estimated, it can be used to guide further tasks such as bin picking, trajectory planning of the robotic arm, or manipulation of the bin and its content using a gripper.

The task of pose estimation involves accurately determining an object's position and orientation in space using data from RGB cameras or 3D scanners. The poses are usually represented using 3D translation and 3D rotation, which together have 6 degrees of freedom. This task has a long history within computer vision research \cite{huttenlocher1990recognizing} and is still actively researched \cite{hodan2024bop}.

In recent years, this task has been dominated by methods utilizing deep learning. Commonly, these methods require training data for individual instances of objects \cite{hodan2024bop}, which limits their reusability when different objects are considered. To address this shortcoming few-shot methods can be utilized \cite{wen2024foundationpose}, but they often still require a CAD model of the specific object instance or several reference images. 

Our paper focuses on a more specific problem of estimating the 6DoF poses of bins. This can be considered a category-level pose estimation problem with various methods available \cite{wang2019normalized,hspose}. In contrast to these methods, which are suited for general categories, we propose a method that leverages the specific geometry of bins used in industrial settings, thus allowing us to achieve superior pose estimation accuracy.

We consider bins to have a cuboid shape with dimensions that can be easily measured in advance. This property of the problem was already exploited by \cite{prepravky} who used it to directly regress the 6DoF pose using a convolutional neural network. In this paper, we propose a different approach which exploits the cuboid shape of the bins by first detecting intermediate 3D line segments of top edges and further processing them to obtain the final poses (see Figure \ref{pipeline}). Our method is based on extending a 2D line segment detector \cite{LeTR} from RGB inputs to 3D line segment detection in structured point cloud data.

To evaluate our approach, we use the dataset provided by \cite{prepravky} and extend it with additional newly collected annotated data, which we will make publicly available. The extended dataset includes synthetically generated training data, which we show improves the final pose estimation accuracy. Using the extended dataset we show that our method significantly outperforms other state-of-the-art 6DoF pose estimation methods \cite{prepravky,hspose,wen2024foundationpose} in terms of pose accuracy.

To summarize, this paper provides several contributions:

\begin{itemize}
\item A new method for 6DoF bin pose estimation based on intermediate detection of 3D line segments.
\item A new dataset for the task of 6DoF bin pose estimation from 3D scans with annotated ground truth poses.
\item Evaluation showing that our method achieves superior pose estimation accuracy when compared to other state-of-the-art methods.
\item The new dataset as well as the code for our method, training and evaluation are available online.\footnote{\url{https://github.com/4zzz/LETR3D}}
\end{itemize}

\section{\uppercase{Related Work}}

\subsection{6DoF Object Pose Estimation}

The problem of 6DoF pose estimation has been extensively studied, with numerous analytical and learning-based approaches available. Several variations of the problem are usually considered~\cite{hodan2024bop}. Poses can be estimated for objects that are known in advance and training data with annotations are available or for objects unseen during training. For unseen objects, their 3D models can be provided in the form of CAD models or meshes in advance. However, some methods do not require prior information about the object at all and can instead utilize video \cite{wen2023bundlesdf} or a few reference images \cite{wen2024foundationpose} to estimate the pose.

An extension of this problem is categorical 3D object pose estimation \cite{wang2019normalized,hspose}, where the task includes estimating object dimensions (width, height, depth) along with their poses.

In this paper, we focus on the specific problem of bin pose estimation from 3D camera scans. This problem was addressed in \cite{prepravky}, where the authors have introduced their own manually annotated dataset of 3D scans of bins. Alongside the dataset, the authors also propose a simple deep convolutional neural network which was shown to achieve better results than other 6DoF pose estimation methods. The task is inspired by practical industrial settings in which it is easy to obtain the base dimensions of bins, but their full CAD models are not readily available.

\subsection{Pose Estimation via Intermediate Detection}

It is possible to use deep neural networks to directly regress the object pose \cite{gao20206d,prepravky,lunayach2024fsd}, however, this comes with potential pitfalls of choosing representation for rotation \cite{zhou:2019:rotation} and object symmetries \cite{pitteri2019object}. To address these issues, pose estimation can instead rely on predicting simpler geometric entities, which are then used to obtain the final pose by utilizing geometric methods \cite{lepetit2009ep}. Various types of intermediate geometric objects can be used for this purpose such as 2D lines \cite{dhome2002determination}, keypoints \cite{tekin2018real}, correspondences \cite{ausserlechner2024zs6d} or fragments \cite{hodan2020epos}. In this paper we propose to use 3D line segments as the intermediate primitives which are then used to obtain the final object pose.

\subsection{Line Segment Detection}

Accurate detection of line segments plays a crucial role in various computer vision and robotics applications, particularly in SLAM \cite{pl_slam}, and relative pose estimation \cite{duff2019plmp}.

To estimate 2D line segments traditional methods such as the Hough Transform \cite{HoughTransfrom} with edge detection \cite{canny2009computational} are widely used for detecting global line structures by voting in a parameterized space. The Line Segment Detector (LSD) \cite{LSD} improves upon this by efficiently detecting local segments without requiring prior edge linking. More recent learning-based approaches, such as DeepLSD \cite{Pautrat_2023_DeepLSD}, leverage deep neural networks to enhance robustness to varying conditions, learning structured representations for accurate segment detection. Transformer-based methods like LeTR \cite{LeTR} further advance the field by modeling long-range dependencies and achieving superior detection in complex scenes.

Detection of 3D line segments is less studied. Several methods for 3D line segment detection leverage multiple-view geometry with known camera positions, which can be automatically reconstructed. By identifying the endpoints of the same line segment across multiple views, 2D line segments can then be back-projected into 3D using triangulation \cite{10.1007/978-3-642-19309-5_31,Liu_2023_LIMAP}.

For point cloud data, an analytical method was developed by \cite{lu2019fast}. To estimate 3D line segments the input point cloud is segmented into parts that lie close to detected planes. Points are then projected onto their corresponding plane and 2D line detection is performed on projected points. Then, 2D line segments are backprojected to 3D. 

A related task to 3D line segment detection is wireframe detection, which, in addition to detecting individual line segments with two endpoints, also identifies junction points where multiple line segments meet. In \cite{liu2021pc2wf} the authors propose a method for wireframe detection on a point cloud. In~\cite{zhou2019learning} only a single RGB is used to estimate wireframes of objects with orthogonal structure.

\section{\uppercase{Bin Pose Estimation}}

In this section we present an overview of our method for 3D line segment detection and its application for the task of 6DoF bin pose estimation. The pipeline of our method is depicted in Figure~\ref{pipeline}.

\subsection{3D Line Segment Estimation Network}
\label{network}
To detect 3D line segments we modify a 2D line segment detection network LeTR~\cite{LeTR}. LeTR follows a simple architecture based on the DeTR object detector~\cite{DeTR}. In LeTR a deep convolutional backbone extracts features, producing two outputs: fine features from an earlier layer and coarse features from the final layer. These features, serialized and enriched with positional encoding, pass through a transformer architecture. The output is processed by two feed-forward networks: one responsible for predicting segment endpoints and the other for producing a one-hot classification vector over two classes (line and no-line), which is subsequently used to derive an estimation confidence using a softmax function.

To adapt this model for 3D, we modify the final stage, enabling the feed-forward network to predict six values corresponding to the two 3D endpoints. Additionally, we remove the sigmoid activation as in the 3D scenario it is not possible to easily normalize the depth coordinate, since its maximal value is unknown. To successfully train LeTR on smaller datasets, it is essential to use the pre-trained weights from the DeTR model.

To train our model, we follow the training procedure proposed in the original LeTR paper. It consists of three stages. In the first stage, only a part of the network is trained using the backbone and one transformer to predict line segments from extracted coarse features. In the second stage, the output of the first transformer is passed to a second transformer along with a fine feature sequence to produce the final prediction. During this stage, the weights trained in the first stage remain frozen. The third stage trains the network by using the focal loss~\cite{focalloss} which helps improve classification for difficult samples.

The loss function is computed following the approach in \cite{LeTR}. The network generates a fixed maximum number of predictions, which are then matched with ground-truth annotations using bipartite matching, optimized based on the matching loss. For classification, a binary cross-entropy loss is applied, while the L1 loss is used to measure the difference between the matched predicted and annotated endpoints. Since the L1 loss operates on points of any dimension, it can be used when extending the network to predict 3D line segments.

\subsection{Bin Pose Estimation}

To estimate the 3D pose from the predicted line segments, we select the four segments with the highest confidence scores. Let $E$ be a matrix whose rows contain the eight endpoints $\vec{p}_i$ of these selected segments centered by their centroid $\vec{p}_c = \frac{1}{8}\sum_{i=1}^8 \vec{p}_i$ as:
\begin{equation}    
E = \begin{bmatrix} 
    \vec{p}_1 - \vec{p}_c\\ 
    \vdots \\ 
    \vec{p}_8 - \vec{p}_c
\end{bmatrix} 
\in \mathbb {R}^{8 \times 3}.
\end{equation}
To determine the bin's orientation, we fit a plane to the endpoints of the selected line segments using Singular Value Decomposition (SVD) of $E = U \Sigma V^T$. The normal vector of the best-fitting plane is given by the right singular vector corresponding to the smallest singular value (last column of $V$). Since we assume the bin's top face is oriented upwards, we ensure that the normal vector $\vec{n}$ points in the positive direction on the z axis. If necessary, we flip its direction.

Additionally, the two longest predicted segments provide two direction vectors, denoted as $\vec{d_1}$ and $\vec{d_2}$. We use them to estimate the bin's direction vector $\vec{d}$ along its longer sides by computing their mean while ensuring they do not oppose each other. To do this, we flip one of the vectors if their dot product is less than 0 as:
\begin{equation}
\vec{d} =
\begin{cases}
    \frac{\vec{d}_1 + \vec{d}_2}{2} \quad \text{if $\vec{d}_1 \cdot \vec{d}_2 \geq 0$,} \\
    \frac{\vec{d}_1 - \vec{d}_2}{2} \quad \text{otherwise}.
\end{cases}
\end{equation}
To obtain the rotation matrix, we perform Gram-Schmidt orthogonalization process on vectors $\vec{n}$ and $\vec{d}$, thus obtaining the columns ($\vec{r_x}, \vec{r_y}, \vec{r_z}$) of the rotation matrix:

\begin{figure}
    \begin{tabular}{cccc} 
        \includegraphics[width=\columnwidth]{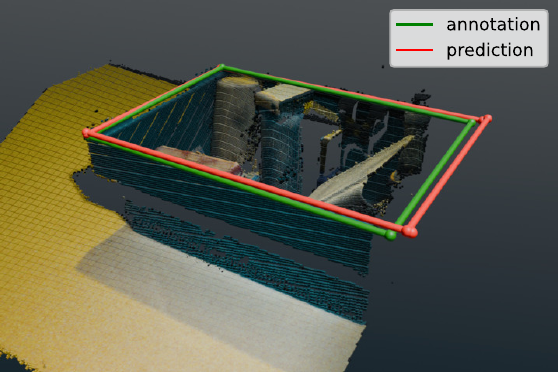} \\
        \includegraphics[width=\columnwidth]{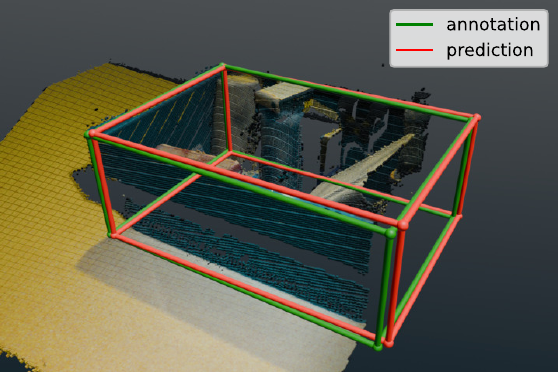} \\
    \end{tabular}
    \caption{\textbf{Top}: 3D line segments detected by our method and the annotated line segments. \textbf{Bottom}: Estimated pose based on the predicted line segments compared with the annotated pose.} 
    \label{diff} 
\end{figure}

\begin{figure*}
\centering
        \includegraphics[width=0.22\textwidth]{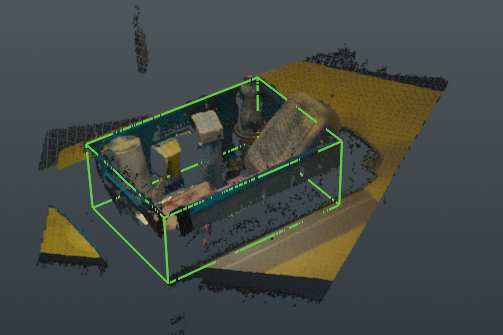} \hfill \includegraphics[width=0.22\textwidth]{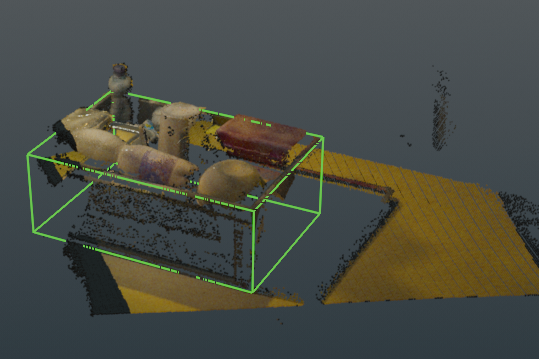} \hfill
        \includegraphics[width=0.22\textwidth]{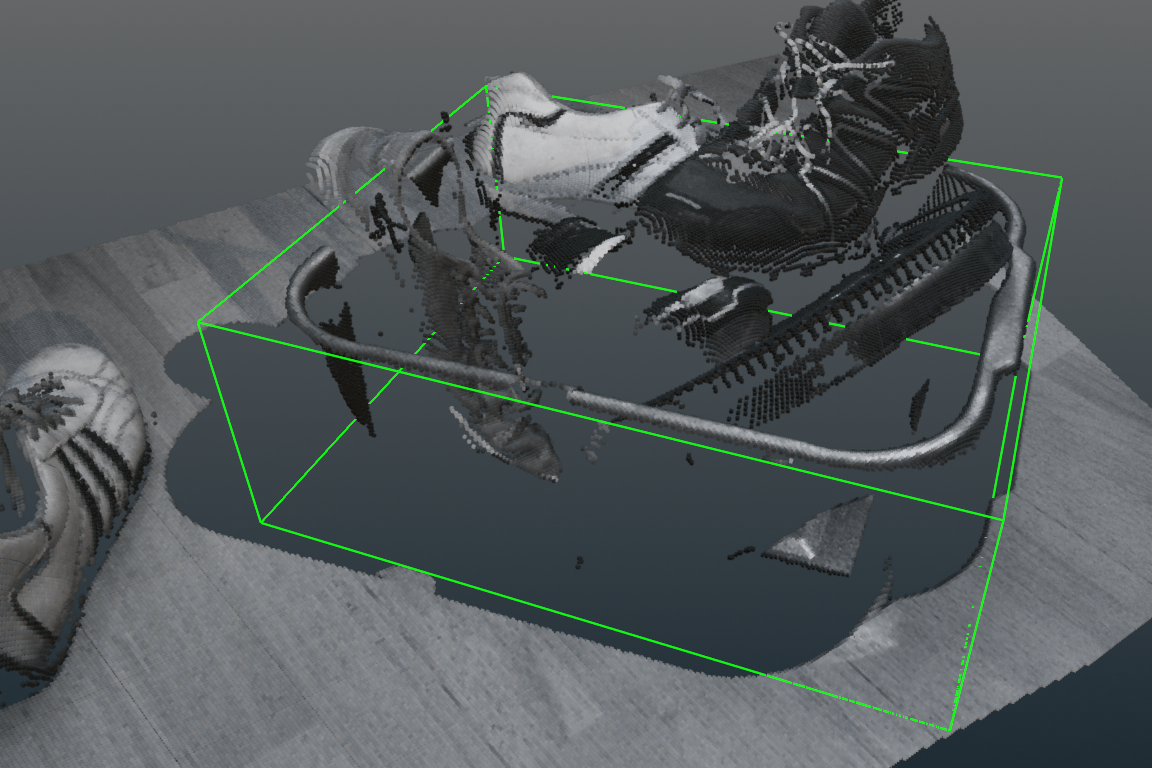} \hfill
        \includegraphics[width=0.22\textwidth]{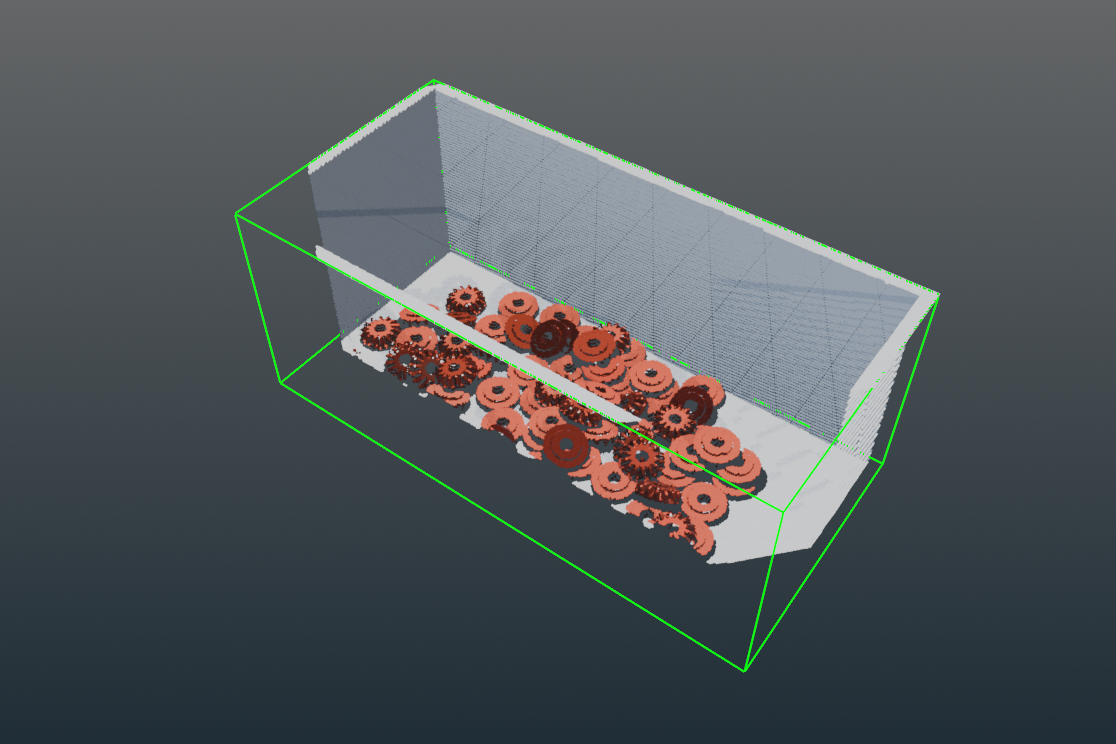}
\caption{Samples from the extended dataset of 3D scans of bins. Green lines indicate the annotated bin positions. The three images on the left show real scans, while the image on the right shows synthetically generated scans.} 
\label{fig:dataset}
\end{figure*} 

\begin{align}
    \vec{r}_z &= \frac{\vec{n}}{||\vec{n}||}, \\
    \hat{\vec{r}}_x &= \vec{d} - \left( \vec{r}_z \cdot \vec{d} \right) \vec{r}_z, \\
    \vec{r}_x &= \frac{\hat{\vec{r}}_x}{||\hat{\vec{r}}_x||}, \\
    \vec{r}_y &= \vec{r}_z \times \vec{r}_x.
\end{align}

To estimate the translation, we could compute the centroid of the endpoints and move in the direction towards the bin’s bottom by a bin-specific amount $d_{bin}$. However, in some cases, the line segment predictor outputs multiple segments close together, which may result in missing segments on other sides of the bin. As a consequence, the computed center of endpoints may be shifted. The endpoints should cluster around four centers that approximate a rectangle in the upper part of the bin. To estimate these centers, we iteratively merge the closest points, replacing them with their mean, until only four points remain. We denote these points as $\vec{q}_i$ for $i \in \left\{1,2,3,4\right\}$.

Then we get our predicted translation $\vec{t}$ vector by using the estimated normal vector pointing upwards $\vec{r}_z$ as:
\begin{equation}    
    \vec{t} = \frac{1}{4} \sum_{i=1}^{4} \vec{q}_i - d_{bin}\vec{r}_z ,
\end{equation}
where $d_{bin}$ is the height of the bin, which is available from the bin model. An example of the estimated bin pose is shown in Figure~\ref{diff}.

\section{\uppercase{Dataset}}

In this section, we provide information on the dataset used for training and evaluation. We use data introduced by~\cite{prepravky} that can be seen in Figure~\ref{fig:dataset} along with a newly collected dataset.

\subsection{New Dataset}

We have captured new data using an automated setup. It was captured using Univeral Robots UR5E robotic arm,\footnote{https://universal-robots.com/products/ur5e/} with two attached 3D cameras. Specifically, we employed industrial-grade Motion Cam S+ cameras from Photoneo,\footnote{https://photoneo.com/products/motioncam-3d-s-plus/} featuring blue laser projectors. One of the cameras was fitted with a custom wide lens, providing a slightly different view of the scene. In total, 10 different scenes with various bin models were captured. Each scene was observed from multiple viewpoints, ranging between 25 and 30 positions. Both cameras were registered into a single world coordinate system using a marker board, and individual frames were subsequently recalculated into their respective camera spaces based on precise transformations from the robotic arm. We make the dataset publicly available.\footnote{Link not provided to preserve blindness.}

\subsection{Data Split}

We combine the newly acquired scans along with the data from \cite{prepravky} into a single dataset and split it into training, validation, and test sets. Our training set contains 1507 scans, including all 657 synthetic samples from \cite{prepravky}. Additionally, we reserve 100 real scans for validation.

To prevent data leakage in the test split, the scans are grouped into scenes, where each scene contains the same bin model and shares other common characteristics, such as the physical environment or additional objects present. We reserve entire scenes exclusively for the test set. This results in the test set consisting of 100 scans. This split is provided along with the dataset to facilitate reproducibility. 

\section{\uppercase{Evaluation}}

In this section, we provide an evaluation of our method for the task of 6DoF bin pose estimation.

\begin{figure*}
\centering
        \includegraphics[width=0.22\textwidth]{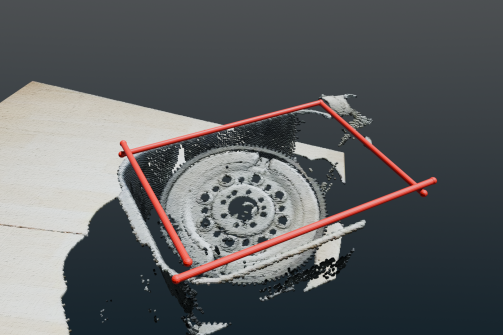} \hfill \includegraphics[width=0.22\textwidth]{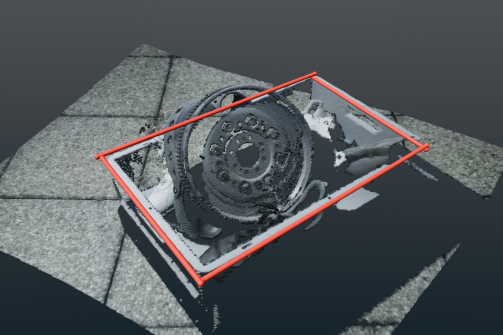} \hfill
        \includegraphics[width=0.22\textwidth]{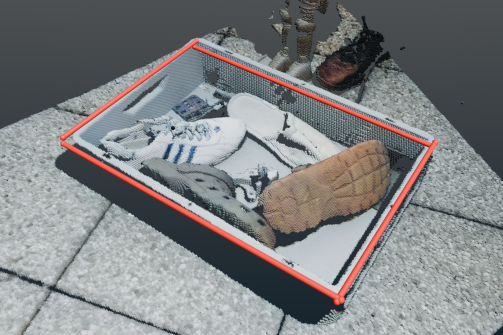} \hfill
        \includegraphics[width=0.22\textwidth]{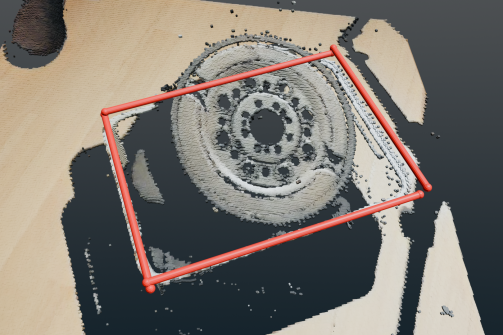}
\caption{Lines predicted by our best model on samples from the test set. The leftmost image depicts a bin scan with a missing corner; despite this, the model is able to infer the position of the missing corner.} 
\label{fig:predictions}
\end{figure*} 

\subsection{Metrics}
For evaluation, we use the mean translation error ($\overline{\textit{e}_{TE}}$) and the mean rotation error ($\overline{\textit{e}_{RE}}$) as defined in~\cite{prepravky}. To calculate the translation error ($\textit{e}_{TE}$) we use the Euclidean distance between the predicted and ground truth translation. For rotation error ($\textit{e}_{RE}$) we compute the angular distance between the predicted ($R$) and ground truth rotation matrix ($\hat{R}$) in degrees using the following formula:
\begin{equation}
    \textit{e}_{RE}(\hat{R}, R) = \arccos\left(\frac{\text{Tr}(\hat{R}R^{-1})-1}{2}\right).
\end{equation}

\subsection{Training}

As described in Section~\ref{network}, we train our models in three stages, using ResNet-50 \cite{ResNet} as the backbone in all cases.
\begin{itemize}
    \item Stage 1: The model is trained for 300 epochs with an initial learning rate of $10^{-4}$, which is reduced by a factor of 10 every 150 epochs.
    \item Stage 2: Training continues for an additional 100 epochs, maintaining the same initial learning rate and applying a learning rate drop by a factor of 10 every 30 epochs.
    \item Stage 3: The final stage consists of 20 epochs with a fixed learning rate of $10^{-5}$.
\end{itemize}

Before training, we normalize our dataset to ensure that the sample channels have zero mean and unit variance.

\subsection{Ablation Study}
To find the best hyperparameter settings, we use the validation set. We performed hyperparameter search for the number of queries and cutout augmentation \cite{Cutout}.

Additionally, we evaluate the contribution of using synthetic data in the training set to pose prediction accuracy.

\begin{table}
\centering
\begin{tabular}{c|c|c}
    Queries & $\overline{\textit{e}_{TE}}$ [cm] & $\overline{\textit{e}_{RE}}$ [$\degree$] \\
    \hline
    4   & \textbf{2.518} & \textbf{3.804} \\
    8   & 2.583 & 4.117 \\
    16  & 2.807 & 4.281 \\
    32  & 2.945 & 5.327 \\
    64  & 2.526 & 3.865 \\
    128 & 2.779 & 5.933 \\
    256 & 9.215 & 21.313 \\
\end{tabular}
\caption{Effect of the number of queries output by the network on validation pose estimation accuracy.}
\label{tbl_queries}
\end{table}

\paragraph{Queries.}  
The number of queries determines how many predictions the model generates and was found in the original paper to influence the accuracy of its predictions. We evaluated several networks with different numbers of queries. The results are presented in Table~\ref{tbl_queries} and show that predicting only 4 line segments leads to the best performance.

\paragraph{Augmentation.}
In this experiment, we apply cutout augmentation \cite{Cutout} to our dataset. Cutout augmentation removes points within a randomly selected rectangular region in the structured point cloud. 

We set the probability of applying cutout $c_p = 0.5$, the minimum cutout area $c_{\min} = 0.2$, and we evaluate the effects of setting the maximum cutout area $c_{\max}$. 

From the results shown in Table~\ref{tbl_cutout}, we conclude that using this augmentation had a negligible effect on mean translation error but significantly improved mean rotation error when $c_{\max}$ is set to 0.8. 

\begin{table}
\centering
\begin{tabular}{c|c|c}
    $c_{\text{max}}$ & $\overline{\textit{e}_{TE}}$ [cm] & $\overline{\textit{e}_{RE}}$ [$\degree$] \\
    \hline
    0.0 & \textbf{2.518} & 4.117 \\
    0.3 & 2.564 & 3.671 \\
    0.4 & 2.879 & 3.863 \\
    0.6 & 2.620 & 3.795 \\
    0.8 & 2.637 & \textbf{3.306} \\
\end{tabular}
\caption{Effect of different settings of the maximum cutout area ($c_{\max}$) on the validation pose estimation accuracy with query hyperparameter set to 4. Setting $c_{\max} = 0$ disables cutout augmentation.}
\label{tbl_cutout}
\end{table}

\paragraph{Synthetic data contribution.}

We train both our model and the convolutional model from \cite{prepravky} on the training set with and without synthetic data to evaluate its effect on pose prediction accuracy. Our results, presented in Table~\ref{tbl_real_and_synth}, show a significant improvement in pose estimation accuracy on the validation subset of our dataset, which contains only real scans. This is in line with the suggestion from \cite{prepravky}, which proposes that training on synthetic data improves the prediction accuracy on real scans.

\begin{table}
\centering
\resizebox{\columnwidth}{!}{
\begin{tabular}{c|c|c|c}
    Model & Train Set &  $\overline{\textit{e}_{TE}}$ [cm] & $\overline{\textit{e}_{RE}}$ [$\degree$] \\
    \hline
    \multirow{2}{*}{CNN baseline} & real+synthetic & \textbf{4.029} & \textbf{10.440} \\
      & real & 4.506 & 11.542 \\
    \hline
    \multirow{2}{*}{LeTR 3D} & real+synthetic & \textbf{2.391} & \textbf{3.949} \\
     & real & 2.719 & 6.418
\end{tabular}
}
\caption{Contribution of synthetic training samples to validation pose estimation accuracy.}
\label{tbl_real_and_synth}
\end{table}

\subsection{Results}

According to the validation results, we train a final model on both real and synthetic data, set the query count to 4, and augment our dataset with cutout using $c_{\max} = 0.8$. Examples of predicted lines on some of our test samples can be seen in Figure~\ref{fig:predictions}.

For comparison, we train a baseline model that directly predicts translation and rotation \cite{prepravky} using the training set of the full dataset. We use the hyperparameters suggested in the paper.

Additionally, we evaluated the accuracy of pose estimation of two additional baselines: a category-level object pose estimation method HS-Pose \cite{hspose} and a zero-shot foundational model for 6DoF pose estimation FoundationPose \cite{wen2024foundationpose}. During training, both models require a segmentation mask of the target object. We derive this mask from the ground-truth pose and the bin’s bounding box for each sample. These models also utilize CAD models during prediction. We do not have precise bin reconstructions or exact CAD models available, therefore we use simplified versions based on the available bin dimensions. We train HS-Pose on our full training set, while for FoundationPose we use pretrained weights, as it supports pose estimation of novel objects without fine-tuning.

We evaluate all models on the test set of our dataset. Results in Table~\ref{tbl_results} show that our method significantly outperforms the previous convolutional model introduced by \cite{prepravky} as well as newer state-of-the-art pose estimation models both in terms of the translation and rotation errors. Moreover, compared to HS-Pose and FoundationPose, our method does not require a CAD model of the bins nor any intermediate segmentation of the bins.

\begin{table}
\centering
\begin{tabular}{c|c|c}
    Model & $\overline{\textit{e}_{TE}}$ [cm] & $\overline{\textit{e}_{RE}}$ [$\degree$] \\
    \hline
    FoundationPose & 5.603 & 15.298 \\
    CNN Baseline & 5.791 & 25.162 \\
    HS-Pose & 3.398 & 20.724 \\
    LeTR 3D (ours) & \textbf{3.053} & \textbf{8.273} \\
\end{tabular}
\caption{Comparison of pose estimation accuracy for trained models on the test set.}
\label{tbl_results}
\end{table}

\section{\uppercase{Conclusion}}

In this paper, we deal with the problem of 6DoF pose estimation of bins from input 3D scans. To facilitate research into this task, we have collected a novel dataset with annotated 3D scans of bins, which we made publicly available. We also propose a novel method for 6DoF bin pose estimation using intermediate detection of 3D line segments. In our experiments, we show the superior pose estimation accuracy of our method compared to other state-of-the-art approaches. In line with previous results, we also show that synthetic training data can significantly improve the performance of deep learning methods when only limited real training data is available. 

\paragraph{Limitations and Future Work}

A limitation of our proposed method is that it requires cuboidal bins with parallel opposite edges. When the bin has oval corners, the line can still fit in the linear parts of the edges, but the bin must be symmetrical, with the linear parts parallel to each other.
Additionally, our method assumes that the bins are oriented upwards in the z-coordinate, so all the 4 detected lines and their 8 corresponding endpoints lie in the top plane of the bin. In order to infer edges lines in the top plane of the bin, the neural model is trained only using the top line annotations. Our method would fail in scenarios with overturned or flipped bins, where the bin is not oriented upwards in the z-coordinate. 
As a future work, voting-based filtering could be used to filter out detected edges outside the top plane. Currently, 4 lines with the highest confidence are taken as the top plane edges.
Moreover, improvements in the synthetic data generation process and its use during training are interesting avenues for future research.

\section*{\uppercase{Acknowledgements}}

This work was funded by the EU NextGenerationEU through the Recovery and Resilience Plan for Slovakia under the project ''InnovAIte Slovakia, Illuminating Pathways for AI-Driven Breakthroughs" No.~09I02-03-V01-00029.

\bibliographystyle{apalike}
{\small
\bibliography{example}}

@ARTICLE{LSD,
  author={Grompone von Gioi, Rafael and Jakubowicz, Jeremie and Morel, Jean-Michel and Randall, Gregory},
  journal={IEEE Transactions on Pattern Analysis and Machine Intelligence}, 
  title={LSD: A Fast Line Segment Detector with a False Detection Control}, 
  year={2010},
  volume={32},
  number={4},
  pages={722-732},
  doi={10.1109/TPAMI.2008.300}}

@InProceedings{Pautrat_2023_DeepLSD,
    author = {Pautrat, Rémi and Barath, Daniel and Larsson, Viktor and Oswald, Martin R. and Pollefeys, Marc},
    title = {DeepLSD: Line Segment Detection and Refinement with Deep Image Gradients},
    booktitle = {Computer Vision and Pattern Recognition (CVPR)},
    year = {2023},
    pages={17327-17336}
}

@InProceedings{LeTR,
	author    = {Xu, Yifan and Xu, Weijian and Cheung, David and Tu, Zhuowen},
	title     = {Line Segment Detection Using Transformers Without Edges},
	booktitle = {CVPR},
	month     = {June},
	year      = {2021},
	pages     = {4257-4266}
}

@InProceedings{Liu_2023_LIMAP,
    author = {Liu, Shaohui and Yu, Yifan and Pautrat, Rémi and Pollefeys, Marc and Larsson, Viktor},
    title = {3D Line Mapping Revisited},
    booktitle = {Computer Vision and Pattern Recognition (CVPR)},
    year = {2023},
    pages={21445-21455}
}

@InProceedings{10.1007/978-3-642-19309-5_31,
author="Chen, Tingwang
and Wang, Qing",
editor="Kimmel, Ron
and Klette, Reinhard
and Sugimoto, Akihiro",
title="3D Line Segment Detection for Unorganized Point Clouds from Multi-view Stereo",
booktitle="Computer Vision -- ACCV 2010",
year="2011",
publisher="Springer Berlin Heidelberg",
address="Berlin, Heidelberg",
pages="400--411",
isbn="978-3-642-19309-5"
}

@article{lu2019fast,
title={Fast 3D Line Segment Detection From Unorganized Point Cloud},
author={Xiaohu, Lu and Yahui, Liu and Kai, Li},
journal={arXiv preprint arXiv:1901.02532},
year={2019},
}

@inproceedings{zhou:2019:rotation,
  title={On the continuity of rotation representations in neural networks},
  author={Zhou, Yi and Barnes, Connelly and Lu, Jingwan and Yang, Jimei and Li, Hao},
  booktitle={Proceedings of the IEEE/CVF Conference on Computer Vision and Pattern Recognition},
  pages={5745--5753},
  year={2019}
}

@inproceedings{prepravky,
  title={Towards Deep Learning-based 6D Bin Pose Estimation in 3D Scans},
  author={Gajdo{\v{s}}ech, Luk{\'a}{\v{s}} and Kocur, Viktor and Stuchl\'{i}k, Martin and Hudec, Luk{\'a}{\v{s}} and Madaras, Martin},
  booktitle={VISAPP},
  pages={545--552},
  year={2022},
  month={February},
  organization={Scitepress}
}

@inproceedings{hodan2024bop,
  title={Bop challenge 2023 on detection segmentation and pose estimation of seen and unseen rigid objects},
  author={Hodan, Tomas and Sundermeyer, Martin and Labbe, Yann and Nguyen, Van Nguyen and Wang, Gu and Brachmann, Eric and Drost, Bertram and Lepetit, Vincent and Rother, Carsten and Matas, Jiri},
  booktitle={Proceedings of the IEEE/CVF Conference on Computer Vision and Pattern Recognition},
  pages={5610--5619},
  year={2024}
}

@inproceedings{focalloss,
  title={Focal loss for dense object detection},
  author={Lin, Tsung-Yi and Goyal, Priya and Girshick, Ross and He, Kaiming and Doll{\'a}r, Piotr},
  booktitle={Proceedings of the IEEE international conference on computer vision},
  pages={2980--2988},
  year={2017}
}

@inproceedings{zhou2019learning,
  title={Learning to reconstruct 3d manhattan wireframes from a single image},
  author={Zhou, Yichao and Qi, Haozhi and Zhai, Yuexiang and Sun, Qi and Chen, Zhili and Wei, Li-Yi and Ma, Yi},
  booktitle={Proceedings of the IEEE/CVF international conference on computer vision},
  pages={7698--7707},
  year={2019}
}

@article{liu2021pc2wf,
  title={Pc2wf: 3d wireframe reconstruction from raw point clouds},
  author={Liu, Yujia and D'Aronco, Stefano and Schindler, Konrad and Wegner, Jan Dirk},
  journal={arXiv preprint arXiv:2103.02766},
  year={2021}
}

@inproceedings{hodan2020epos,
  title={Epos: Estimating 6d pose of objects with symmetries},
  author={Hodan, Tomas and Barath, Daniel and Matas, Jiri},
  booktitle={Proceedings of the IEEE/CVF conference on computer vision and pattern recognition},
  pages={11703--11712},
  year={2020}
}

@inproceedings{DeTR,
  title={End-to-end object detection with transformers},
  author={Carion, Nicolas and Massa, Francisco and Synnaeve, Gabriel and Usunier, Nicolas and Kirillov, Alexander and Zagoruyko, Sergey},
  booktitle={European conference on computer vision},
  pages={213--229},
  year={2020},
  organization={Springer}
}

@INPROCEEDINGS{ResNet,
  author={He, Kaiming and Zhang, Xiangyu and Ren, Shaoqing and Sun, Jian},
  booktitle={2016 IEEE Conference on Computer Vision and Pattern Recognition (CVPR)}, 
  title={Deep Residual Learning for Image Recognition}, 
  year={2016},
  volume={},
  number={},
  pages={770-778},
  doi={10.1109/CVPR.2016.90}}

@article{cutout,
  title={Improved regularization of convolutional neural networks with cutout},
  author={DeVries, Terrance and Taylor, Graham W},
  journal={arXiv preprint arXiv:1708.04552},
  year={2017}
}

@article{HoughTransfrom,
  title={Use of the Hough Transformation to Detect Lines and Curves in Pictures},
  author={Duda, Richard O. and Hart, Peter E.},
  journal={Communications of the ACM},
  volume={15},
  number={1},
  pages={11--15},
  year={1972},
  publisher={ACM},
  doi={10.1145/361237.361242}
}

@INPROCEEDINGS{pl_slam,
  author={Pumarola, Albert and Vakhitov, Alexander and Agudo, Antonio and Sanfeliu, Alberto and Moreno-Noguer, Francese},
  booktitle={2017 IEEE International Conference on Robotics and Automation (ICRA)}, 
  title={PL-SLAM: Real-time monocular visual SLAM with points and lines}, 
  year={2017},
  volume={},
  number={},
  pages={4503-4508},
  doi={10.1109/ICRA.2017.7989522}}

@inproceedings{wen2023bundlesdf,
  title={Bundlesdf: Neural 6-dof tracking and 3d reconstruction of unknown objects},
  author={Wen, Bowen and Tremblay, Jonathan and Blukis, Valts and Tyree, Stephen and M{\"u}ller, Thomas and Evans, Alex and Fox, Dieter and Kautz, Jan and Birchfield, Stan},
  booktitle={Proceedings of the IEEE/CVF Conference on Computer Vision and Pattern Recognition},
  pages={606--617},
  year={2023}
}

@inproceedings{wen2024foundationpose,
  title={Foundationpose: Unified 6d pose estimation and tracking of novel objects},
  author={Wen, Bowen and Yang, Wei and Kautz, Jan and Birchfield, Stan},
  booktitle={Proceedings of the IEEE/CVF Conference on Computer Vision and Pattern Recognition},
  pages={17868--17879},
  year={2024}
}

@inproceedings{gao20206d,
  title={6d object pose regression via supervised learning on point clouds},
  author={Gao, Ge and Lauri, Mikko and Wang, Yulong and Hu, Xiaolin and Zhang, Jianwei and Frintrop, Simone},
  booktitle={2020 IEEE International Conference on Robotics and Automation (ICRA)},
  pages={3643--3649},
  year={2020},
  organization={IEEE}
}

@inproceedings{wang2019normalized,
  title={Normalized object coordinate space for category-level 6d object pose and size estimation},
  author={Wang, He and Sridhar, Srinath and Huang, Jingwei and Valentin, Julien and Song, Shuran and Guibas, Leonidas J},
  booktitle={Proceedings of the IEEE/CVF conference on computer vision and pattern recognition},
  pages={2642--2651},
  year={2019}
}

@inproceedings{lunayach2024fsd,
  title={Fsd: Fast self-supervised single rgb-d to categorical 3d objects},
  author={Lunayach, Mayank and Zakharov, Sergey and Chen, Dian and Ambrus, Rares and Kira, Zsolt and Irshad, Muhammad Zubair},
  booktitle={2024 IEEE International Conference on Robotics and Automation (ICRA)},
  pages={14630--14637},
  year={2024},
  organization={IEEE}
}

@inproceedings{pitteri2019object,
  title={On object symmetries and 6d pose estimation from images},
  author={Pitteri, Giorgia and Ramamonjisoa, Micha{\"e}l and Ilic, Slobodan and Lepetit, Vincent},
  booktitle={2019 International conference on 3D vision (3DV)},
  pages={614--622},
  year={2019},
  organization={IEEE}
}

@inproceedings{tekin2018real,
  title={Real-time seamless single shot 6d object pose prediction},
  author={Tekin, Bugra and Sinha, Sudipta N and Fua, Pascal},
  booktitle={Proceedings of the IEEE conference on computer vision and pattern recognition},
  pages={292--301},
  year={2018}
}

@article{lepetit2009ep,
  title={EP n P: An accurate O (n) solution to the P n P problem},
  author={Lepetit, Vincent and Moreno-Noguer, Francesc and Fua, Pascal},
  journal={International journal of computer vision},
  volume={81},
  number={2},
  pages={155--166},
  year={2009},
  publisher={Springer}
}

@inproceedings{ausserlechner2024zs6d,
  title={Zs6d: Zero-shot 6d object pose estimation using vision transformers},
  author={Ausserlechner, Philipp and Haberger, David and Thalhammer, Stefan and Weibel, Jean-Baptiste and Vincze, Markus},
  booktitle={2024 IEEE International Conference on Robotics and Automation (ICRA)},
  pages={463--469},
  year={2024},
  organization={IEEE}
}

@inproceedings{duff2019plmp,
  title={PLMP-point-line minimal problems in complete multi-view visibility},
  author={Duff, Timothy and Kohn, Kathlen and Leykin, Anton and Pajdla, Tomas},
  booktitle={Proceedings of the IEEE/CVF International Conference on Computer Vision},
  pages={1675--1684},
  year={2019}
}

@InProceedings{hspose,
  author    = {Zheng, Linfang and Wang, Chen and Sun, Yinghan and Dasgupta, Esha and Chen, Hua and Leonardis, Ale\v{s} and Zhang, Wei and Chang, Hyung Jin},
  title     = {HS‑Pose: Hybrid Scope Feature Extraction for Category‑Level Object Pose Estimation},
  booktitle = {Proceedings of the IEEE/CVF Conference on Computer Vision and Pattern Recognition (CVPR)},
  month     = {June},
  year      = {2023},
  pages     = {17163--17173},
  doi       = {10.1109/CVPR52729.2023.01646}
}

@article{huttenlocher1990recognizing,
  title={Recognizing solid objects by alignment with an image},
  author={Huttenlocher, Daniel P and Ullman, Shimon},
  journal={International journal of computer vision},
  volume={5},
  number={2},
  pages={195--212},
  year={1990},
  publisher={Springer}
}

@article{canny2009computational,
  title={A computational approach to edge detection},
  author={Canny, John},
  journal={IEEE Transactions on pattern analysis and machine intelligence},
  number={6},
  pages={679--698},
  year={1986},
  publisher={Ieee}
}

@article{dhome2002determination,
  title={Determination of the attitude of 3D objects from a single perspective view},
  author={Dhome, Michel and Richetin, Marc and Lapreste, J-T and Rives, Gerard},
  journal={IEEE transactions on pattern analysis and machine intelligence},
  volume={11},
  number={12},
  pages={1265--1278},
  year={1989},
  publisher={IEEE}
}

\end{document}